\documentclass[sigconf,natbib]{acmart}
\usepackage{tikz}
\usepackage{multirow}
\usetikzlibrary{calc}
\usepackage{graphicx}
\usepackage{pgfplots}  
\usepackage{pgfplots}
\pgfplotsset{compat=1.18}
\usepackage{pgfplotstable}
\usepackage{caption}
\usepackage{subcaption}
\usepackage{tikz}
\usepackage{multirow}
\usetikzlibrary{calc}
\usepackage{graphicx}
\usepackage{pgfplots}  
 \usetikzlibrary{patterns}
\usepackage{pgfplots}
\usepackage{subcaption}
\usepackage{caption}
\pgfplotsset{compat=1.18}
\usetikzlibrary{pgfplots.groupplots}

\AtBeginDocument{%
  \providecommand\BibTeX{{%
    \normalfont B\kern-0.5em{\scshape i\kern-0.25em b}\kern-0.8em\TeX}}}

\setcopyright{acmlicensed}
\copyrightyear{2025}
\acmYear{2025}
\acmDOI{10.1145/3746252.3760998}

\acmConference[CIKM '25]{Proceedings of the 34th ACM International Conference on Information and Knowledge Management}{November 10--14, 2025}{Seoul, Republic of Korea}
\acmBooktitle {Proceedings of the 34th ACM International Conference on Information and Knowledge Management (CIKM 25), November 10--14, 2025, Seoul, Republic of Korea}
\acmISBN{979-8-4007-2040-6/2025/11}

\begin{document}
\newcommand{\customsize}{\fontsize{8.9}{8.9}\selectfont}
\newcommand{\petit}{\fontsize{7.9}{7.9}\selectfont}

\title{Dynamic Triangulation-Based Graph Rewiring for Graph Neural Networks}



\author{Hugo Attali}
\orcid{0009-0003-6887-9146}
\affiliation{%
  \institution{LIPN, Université Sorbonne Paris Nord}
  \city{Villetaneuse}
  \country{France}
}
\email{attali@lipn.univ-paris13.fr}

\author{Thomas Papastergiou}
\orcid{0000-0002-0241-080X}
\affiliation{%
  \institution{LIPN, Université Sorbonne Paris Nord}
  \city{Villetaneuse}
  \country{France}
}
\email{papastergiou@lipn.univ-paris13.fr}

\author{Nathalie Pernelle}
\orcid{0000-0003-1487-393X}
\affiliation{%
  \institution{LIPN, Université Sorbonne Paris Nord}
  \city{Villetaneuse}
  \country{France}
}
\email{pernelle@lipn.univ-paris13.fr}

\author{Fragkiskos D. Malliaros}
\orcid{0000-0002-8770-3969}
\affiliation{%
  \institution{Université Paris-Saclay, CentraleSupélec, Inria}
  \city{Gif-sur-Yvette}
  \country{France}
  }
\email{fragkiskos.malliaros@centralesupelec.fr}

\renewcommand{\shortauthors}{Attali, et al.}

\begin{abstract}

Graph Neural Networks (GNNs) have emerged as the leading paradigm for learning over graph-structured data. However, their performance is limited by issues inherent to graph topology, most notably oversquashing and oversmoothing. Recent advances in graph rewiring aim to mitigate these limitations by modifying the graph topology to promote more effective information propagation. In this work, we introduce TRIGON, a novel framework that constructs enriched, non-planar triangulations by learning to select relevant triangles from multiple graph views. By jointly optimizing triangle selection and downstream classification performance, our method produces a rewired graph with markedly improved structural properties such as reduced diameter, increased spectral gap, and lower effective resistance compared to existing rewiring methods. Empirical results demonstrate that TRIGON outperforms state-of-the-art approaches on node classification tasks across a range of homophilic and heterophilic benchmarks.

\end{abstract}

 



\maketitle
\section{Introduction}

Graph Neural Networks (GNNs) \cite{gori2005new,goller1996learning,gilmer2017neural} have become central to learning on structured data, thanks to their ability to propagate and aggregate information across edges. By iteratively exchanging messages between neighboring nodes, GNNs capture local patterns and encode them into expressive representations, enabling strong performance in various domains such as molecular property prediction \cite{survey-geometric-GNNs-2023, pmlr-v202-duval23a}, social network analysis \cite{panagopoulos:hal-04824022}, recommender systems \cite{wu2020comprehensive}, and spatiotemporal forecasting \cite{castro-correa-tnnls24}.
However, the strictly local nature of message passing can be poorly suited to the structure of many real-world graphs. 
This limitation becomes particularly salient in low-homophily graphs \cite{luan2021heterophily,platonov2023critical,wang2024understanding}, where neighboring nodes frequently belong to different classes. In such contexts, the feature distributions within local neighborhoods often exhibit high inter-class variance, violating the underlying assumptions of local homophily exploited by standard message passing schemes.
As a result, aggregating features from neighbors with different labels often introduces irrelevant or misleading information. This degrades the quality of node representations and makes it harder to distinguish between classes, especially when useful information lies beyond the local neighborhood \cite{zheng2022graph,gong2024survey}.

Additionally, structural patterns such as long paths, bottlenecks, or densely connected clusters can obstruct the diffusion of information, leading to a mismatch between the graph’s topology and the model’s capacity to capture relevant information \cite{UNDERSTANDING_bottlenecks,BORF}. 
Consequently, GNNs often fail to effectively model global dependencies, which can limit their performance on graphs with complex or irregular structures \cite{dwivedi2022long}.
To mitigate these issues, recent advances have explored structural modifications, known as graph rewiring \cite{attali2024rewiring}, to alleviate the topological constraints that hinder effective message passing. These techniques seek to reduce bottlenecks and enhance connectivity by selectively adding or removing edges, using either local descriptors such as curvature \cite{UNDERSTANDING_bottlenecks,giraldo2023trade,BORF,fesser2023mitigating,attali2025curvature} or global criteria like spectral expansion\cite{banerjee2022oversquashing,FOSR}.

Another promising strategy, borrowed from computational geometry, is Delaunay triangulation \cite{edelsbrunner2000triangulations,sharp2020pointtrinet}. Traditionally employed for surface meshing and point-cloud analysis, it produces well-conditioned meshes by maximizing each triangle’s minimum angle \cite{huang2024surface}. Applied to graph rewiring, Delaunay triangulation in feature space rebuilds the graph by connecting every triplet of nodes that defines a Delaunay triangle. However, by relying solely on local triangles in feature space, Delaunay triangulation fails to capture non-local or higher-order structures that may be essential for effective downstream classification.


To address this limitation, we propose TRIGON, a graph rewiring framework that formulates triangle selection as an optimization problem. Each triangle is embedded via a learnable encoder and selected through a differentiable  mechanism guided by classification supervision and structural regularization. This approach enables the construction of non-planar graphs with improved local and global structural properties, thereby enhancing the effectiveness of message passing in both homophilic and heterophilic regimes. In summary, our key contributions are summarized as follows:
\begin{enumerate}

\item \textbf{Theoretical analysis of triangle-based rewiring.} We analyze in different ways the effect of triangles on the structure of the graph and their impact on message passing.

\item \textbf{Differentiable triangle selection for graph rewiring.} Based on this analysis, we develop TRIGON, a learnable module that scores and selects triangles based on their relevance to the classification objective. The triangle selection and the GNN are trained jointly through loss functions that encourage the construction of a rewired graph with coherent, balanced, and task-aligned connectivity. This joint optimization leads to more effective message passing and improved downstream performance.

\item \textbf{Comprehensive empirical validation.} We conduct extensive experiments across a set of heterophilic and homophilic benchmark graphs, demonstrating that our learnable rewiring method outperforms both static Delaunay-based and other state-of-the-art rewiring strategies.
\end{enumerate}

\noindent {\textbf{Reproducibility.}} Our code is available\footnote{https://github.com/Hugo-Attali/TRIGON-CIKM-2025}.

\section{Background and Related Work}

We start by introducing notations used throughout this paper. We denote a graph as \(G = (\mathcal{V}, \mathcal{E})\), where \(\mathcal{V}\) is the set of \(N = |\mathcal{V}|\) nodes and \(\mathcal{E} \subseteq \mathcal{V} \times \mathcal{V}\) the set of edges. We consider undirected graphs, i.e., \((i,j) \in \mathcal{E} \Rightarrow (j,i) \in \mathcal{E}\). The adjacency matrix \(\mathbf{A} \in \mathbb{R}^{N \times N}\) is defined by \(\mathbf{A}_{ij} = 1\) if \((i,j) \in \mathcal{E}\), and \(0\) otherwise. The degree of node \(i\) is \(d_i = \sum_j \mathbf{A}_{ij}\), and the diagonal degree matrix is \(\mathbf{D}\) with \(\mathbf{D}_{ii} = d_i\). We also define the \emph{normalized Laplacian} as $\mathbf{L}_{\text{norm}} = \mathbf{I} - \mathbf{D}^{-1/2} \mathbf{A} \mathbf{D}^{-1/2}$. The normalized Laplacian is symmetric positive semi-definite, with eigenvalues \(0 = \lambda_1 \le \lambda_2 \le \dots \le \lambda_N \le 2\).

\subsection{Message Passing in GNNs}

Graph Neural Networks (GNNs) are based on the message passing framework, wherein node representations are iteratively refined through local interactions \cite{gilmer2017neural}. At each layer, a node aggregates information from its immediate neighbors, typically using a permutation-invariant function, followed by a transformation via a learnable mapping. Formally, for a node \(i \in \mathcal{V}\), its representation at layer \(k+1\) is computed as:
\[
\mathbf{h}_i^{(k+1)} = \phi\left(\mathbf{h}_i^{(k)}, \bigoplus_{j \in \mathcal{N}(i)} \psi(\mathbf{h}_j^{(k)})\right),
\]
where \(\mathbf{h}_i^{(k)}\) denotes the representation of node \(i\) at layer \(k\), \(\mathcal{N}(i)\) its neighborhood, \(\psi\) the message function, and \(\phi\) the update function. The operator \(\bigoplus\) denotes a permutation-invariant aggregation function such as sum, mean, or maximum. This iterative process enables GNNs to capture both feature and topological information from graph neighborhoods \cite{gcn,gat}. Message passing is particularly effective on homophilic graphs, where neighboring nodes tend to share similar labels. However, in heterophilic graphs \cite{Homophily} or when long-range dependencies \cite{alon2020bottleneck} are present, information must travel across multiple hops to capture relevant node interactions. Specifically, to allow communication between two nodes at a distance \(k\), at least \(k\) layers of message passing are required. However, increasing the depth introduces two major challenges, namely \emph{oversmoothing} and \emph{oversquashing}.

\paragraph{\textbf{Oversmoothing.}}  Oversmoothing refers to the phenomenon by which node representations become indistinguishable as the depth of a GNN increases \cite{oversmoothing,Over-Smoothing_gnn,GNN_lose} . At each layer, information is propagated and aggregated across neighboring nodes, which gradually reduces the variance between node features. In the limit, this process drives all node embeddings toward a dimensional subspace where they become nearly identical. Consequently, the model's capacity to discriminate between different structural roles or feature patterns deteriorates, often resulting in degraded performance in node-level prediction tasks.
\paragraph{\textbf{Oversquashing.}}
Oversquashing occurs when information from exponentially large neighborhoods is compressed into fixed-size node embeddings through a limited number of message passing steps \cite{alon2020bottleneck}. As the receptive field expands with depth, the aggregation function must encode a growing amount of information into a fixed-dimensional space. This creates a bottleneck that limits the model's ability to capture long-range dependencies, particularly in graphs with complex or sparse topologies.

Importantly, these limitations are not solely determined by network depth \cite{UNDERSTANDING_bottlenecks,di2023over}. They can also be significantly amplified by particular structural patterns within the input graph. For instance,  sparsely connected regions or long chains can intensify oversquashing, while densely connected subgraphs such as cliques can accelerate oversmoothing by amplifying redundant information flow.

To mitigate these effects, recent approaches have proposed modifying the input graph structure through techniques such as rewiring or augmenting connectivity, with the goal of facilitating more efficient message passing and alleviating the impact of harmful substructures \cite{UNDERSTANDING_bottlenecks,BORF, giraldo2023trade,FOSR, fesser2023mitigating,banerjee2022oversquashing,attali2025curvature}.

\subsection{Cheeger Constant and Spectral Graph Properties}

\paragraph{Cheeger constant.}
Let \(G = (\mathcal{V}, \mathcal{E})\) be a finite, undirected graph. For any non-empty subset \( \mathcal{S} \subset \mathcal{V} \) with \( |\mathcal{S}| \le |\mathcal{V}|/2 \), define its edge boundary and volume respectively as \cite{cheeger,cheeger2015lower}:
\[
\partial S = \{\{i,j\}\in \mathcal{E} : i\in \mathcal{S},\;j\in \mathcal{V}\setminus \mathcal{S}\}, 
\quad 
\mathrm{vol}(\mathcal{S})=\sum_{j\in \mathcal{S}}\deg(j).
\]
The \emph{Cheeger constant} of \( G \) is defined as:
\[
h(G)=\min_{\substack{\mathcal{S}\subset \mathcal{V}\\0<|\mathcal{S}|\le|\mathcal{V}|/2}}\frac{|\partial \mathcal{S}|}{\mathrm{vol}(\mathcal{S})}.
\]
A small Cheeger constant signals the presence of bottlenecks, facilitating easy graph partitioning. The Cheeger inequality connects this constant to the spectral gap \(\lambda_2\) of the normalized Laplacian:
\[
\frac{h(G)^2}{2}\ \le\ \lambda_2 \ \le\ 2\,h(G).
\]
Thus, large values of \(h(G)\) imply strong connectivity and efficient mixing of random walks.

\paragraph{Diameter bound.}
For a graph with minimum degree \(\ d_{\min}\) and total volume \(\mathrm{vol}(V)\), the diameter satisfies the following bound~\cite{chung1997spectral}:
\[
\mathrm{diam}(G)\;\le\;\frac{2\ln\bigl(\mathrm{vol}(\mathcal{V})\bigr)}{\ln\bigl(1+\frac{h(G)}{d_{\min}}\bigr)},
\]
which implies that:
\begin{equation}
h(G) \geq d_{\min} \left( \operatorname{vol}(\mathcal{V})^{2 / \operatorname{diam}(G)} - 1 \right).
\label{diam}
\end{equation}
Graphs that simultaneously exhibit sparse edge density, high connectivity, and small diameter are called \emph{expanders}, making them particularly effective for message passing \cite{deac2022expander}.

\subsection{Graph Curvature and Effective Resistance}

Similar to manifolds, curvature serves as a useful concept for characterizing the local structure of a graph.
Several discrete notions of curvature, inspired by Riemannian geometry, have been proposed to capture geometric properties of graphs \cite{Forman_curvature,Olivier_curvature,samal2018comparative,jost2014ollivier}.
These measures vary in the structural features they emphasize, such as vertex degrees, small cycles, and probabilistic neighborhood distributions. 
Intuitively, graph curvature quantifies how the local connectivity around a node or edge deviates from that of a flat (e.g., Euclidean or tree-like) structure, highlighting irregularities in local geometry.
Graph curvature offers a local measure for identifying structural bottlenecks in graphs. As demonstrated in \cite{UNDERSTANDING_bottlenecks}, bottleneck structures are associated with edges that exhibit strongly negative curvature. By contrast, the effective resistance between two nodes provides a global measure of the difficulty of information flow between them.  It aggregates contributions from all paths in the graph, thereby capturing the cumulative influence of the entire topology rather than just local neighborhood structure \cite{resit,chandra1989electrical}. Recent work shows that a high effective resistance between two nodes implies that either a scarcity of connecting paths or that the available paths are relatively long, indicating potential communication bottlenecks \cite{resit}.

\subsection{Graph Rewiring}
Graph rewiring seeks to enhance message passing by altering the graph structure, typically through the addition or deletion of edges. By introducing shortcuts or reinforcing weakly connected regions, rewiring can reduce the distance between relevant nodes, improving information propagation. The selection of edges to modify is guided by various structural metrics. Local methods \cite{UNDERSTANDING_bottlenecks,BORF,giraldo2023trade,fesser2023mitigating,attali2025curvature} employ geometric descriptors such as discrete Ricci curvature \cite{Forman_curvature,Olivier_curvature,samal2018comparative,UNDERSTANDING_bottlenecks} to identify and rewire around structural bottlenecks. These approaches typically add edges near highly negatively curved regions, which are indicative of constrained information flow and potential oversquashing. Global methods \cite{banerjee2022oversquashing,FOSR,resit,barbero2024localityaware} aim to alleviate large-scale topological bottlenecks by using metrics that reflect overall graph connectivity. One such metric is effective resistance\cite{chandra1989electrical,resit}, which quantifies the difficulty of information flow between node pairs. Rewiring strategies like GTR\cite{resit} add edges to minimize effective resistance, thus improving long-range communication.
Spectral methods instead focus on maximizing the spectral gap or the Cheeger constant, which promotes better expansion properties and more efficient information diffusion\cite{banerjee2022oversquashing,FOSR}.
 
Although rewiring is primarily intended to alleviate oversquashing, it can also help reduce oversmoothing by sparsifying overly dense substructures \cite{giraldo2023trade,BORF,attali2024delaunay}. By weakening dense regions such as cliques, where excessive local aggregation tends to homogenize node features, rewiring limits redundant message propagation and preserves representational diversity across layers.
More recently, JDR \cite{linkerhagner2025joint} not only rewires the graph but also denoises the node features, in order to construct a topology that maximizes the spectral alignment between the graph structure and the node features.

Two types of rewiring methods can be distinguished, rewiring methods that modify the upstream graph structure before learning \cite{banerjee2022oversquashing,UNDERSTANDING_bottlenecks,resit,barbero2024localityaware,attali2024delaunay,attali2025curvature} and rewiring methods that modify the upstream graph structure during learning \cite{giraldo2023trade,gutteridge2023drew,qian2023probabilistically}.

Another class of approaches discards the original graph structure entirely, constructing a new one via Delaunay triangulation over the node features. This procedure simultaneously avoids edges with highly negative curvature, indicative of structural bottlenecks, and guarantees that the maximum clique size is three, thereby limiting the risk of oversmoothing due to excessively dense subgraphs \cite{attali2024delaunay}.

Although Delaunay rewiring introduces well structured local connectivity and has been shown to reduce phenomena such as oversquashing and oversmoothing, it inherently limits the scope of rewiring to spatially proximal nodes. This constraint overlooks the potential contribution of non-local triangles, which can play a critical role in enhancing global information flow and improving long-range message propagation. To address this limitation, we will shortly introduce TRIGON, a learnable triangle selection mechanism that favors structures beneficial to message passing.

\section{Why Triangle-Aware Graph Rewiring?}



Recent advances in graph rewiring have highlighted the importance of leveraging local higher-order structures, particularly triangles, to enhance message passing \cite{attali2024delaunay}. While higher-order structures are known to increase the expressive power of GNNs \cite{morris2019weisfeiler}, our approach, which will be detailed in Section \ref{sec:method}, differs in that it does not alter the message passing scheme itself. Instead, we use triangle-based motifs to rewire the graph topology, thereby improving the pathways through which information propagates. Triangles not only encode strong local cohesion but also promote \textit{global connectivity}. Notably, the \textit{effective resistance} of an edge $(i, j)$ admits the upper bound \cite{sotiropoulos2021triangle}:
\begin{equation}
{R_{\text{\textit{eff}}}}(i,j) \leq \frac{2}{t(i,j) + 2} ,
\end{equation}
where $t(i,j)$ denotes the number of triangles containing the edge $(i,j)$. This inverse relationship suggests that edges embedded in many triangles support low-resistance communication pathways. Hence, triangle-based rewiring strategies can enhance both local message-passing effectiveness and global information flow in GNNs. By selecting structurally informative triangles, one can alleviate oversquashing, reduce the effective graph diameter, and improve representation learning, particularly in heterophilous settings where long-range dependencies are essential.

Delaunay rewiring \cite{attali2024delaunay} demonstrates that a triangle-based graph rewiring is useful: by keeping edge curvatures close to zero, it helps mitigate both oversmoothing and oversquashing. In the next paragraph, we argue that Delaunay triangulation, being strictly planar, lacks sufficient complexity to capture long-range interactions.

\subsection{Motivation for a More Expressive Graph Triangulation}

\paragraph{\textbf{Spectral analysis}}
By construction, a Delaunay graph is planar, so its number of edges is \(\mathcal{O}(N)\), and its diameter \(\text{diam}(G)\) satisfies the following lower bound \cite{lipton1979separator}:
\begin{equation}
\text{diam}(G) = \Omega(\sqrt{N}),
\end{equation}
\noindent where the notation $\Omega(\cdot)$ denotes an asymptotic lower bound: writing $\text{diam}(G) \in \Omega(N)$ indicates that the graph diameter grows at least proportionally to $N$ as the number of nodes $N$ increases. In the context of Delaunay graphs, this implies that long-range communication requires many hops, which severely limits the efficiency of message passing in large graphs. Under a bounded-degree assumption, such a high diameter necessarily leads to a small Cheeger constant $h(G)$, as expressed by Eq.~\eqref{diam}. This, in turn, results in poor global connectivity. Equivalently, the spectral gap $\lambda_2$, the second-smallest eigenvalue of the normalized Laplacian matrix, tends to zero in large planar graphs~\cite{louder2012diameter}, reflecting weak expansion properties. These inherent spectral limitations make it difficult to model long-distance dependencies using planar triangulations, thus constraining the expressive power of GNNs operating on such graphs.

\paragraph{\textbf{Curvature analysis}}

In this part, we examine how local and non‑local cycles affect discrete curvature and hence information flow. Let \(G = (\mathcal{V}, \mathcal{E})\)  have maximum degree \(d_{\max}\) and minimum degree \(d_{\min}\).  For each edge \((i,j)\in \mathcal{E}\), let \(\deg(i), \deg(j)\) be its endpoint degrees, $t(i,j)$ be the number of triangles containing \((i,j)\), $\Gamma_{\max}(i,j)$ the maximum number of 4-cycles based at edge $e_{ij}$, and $\gamma_i$ be the count of 4-cycles at $e_{ij}$ without diagonals. Then, the balanced Forman curvature \cite{UNDERSTANDING_bottlenecks} is defined as:
\begingroup
\small 
\begin{equation}
\begin{split}
/
c_{ij}
&= \frac{2}{\deg(i)} + \frac{2}{\deg(j)} - 2 \\
&\quad \underbrace{ + 2\,\frac{t(i,j)}{\max\{\deg(i), \deg(j)\}}
+ \frac{t(i,j)}{\min\{\deg(i), \deg(j)\}}}_{\substack{\text{triangle-induced contribution}\\\text{}}}\\
&\quad \underbrace{+ \frac{(\Gamma_{\max}(i,j))^{-1}}{\max\{\deg(i), \deg(j)\}}\,(\gamma_{i} + \gamma_{j})}
_{\substack{\text{4-cycle-induced contribution}\\\text{}}}.
\end{split}
\end{equation}
\endgroup
In a planar triangulation, such as one induced by Delaunay rewiring \cite{attali2024delaunay}, every face is bounded by exactly three edges. As a result, such graphs cannot contain chordless 4-cycles, i.e., cycles of length four that are not completed into cliques, since any quadruplet of nodes is either not connected or forms chords that break the 4-cycle.
This structural restriction has geometric consequences: curvature-based quantities, such as the Balanced Forman-Ricci curvature \cite{UNDERSTANDING_bottlenecks}, depend on contributions from both triangles and 4-cycles. In planar triangulations, where chordless 4-cycles do not occur, the 4-cycle-induced term vanishes, leaving the curvature entirely determined by triangle contributions and node degrees.
In contrast, non-planar triangulations, which we aim to produce with our TRIGON method, naturally allow chordless 4-cycles to emerge. These contribute positively to the curvature and help reduce the negative values often concentrated around bottlenecks.
Therefore, allowing such motifs enhances the graph’s structural richness, contributes to increased curvature in critical regions, and supports more effective message propagation. Ultimately, meaningful improvements in global connectivity require the inclusion of nonplanar and higher-order motifs. In particular, selecting triangles that link distant or otherwise unconnected regions introduces both chords and chordless 4-cycles, thereby boosting discrete curvature and increasing the spectral gap. 

In the following section, we present our dynamic triangulation rewiring framework (TRIGON), which learns enriched, non-planar graphs by selecting both local and non-local triangles. This leads to structures that better support effective message passing in GNNs.

\begin{figure*}[h]
     \centering
     \includegraphics[width=.95\textwidth]{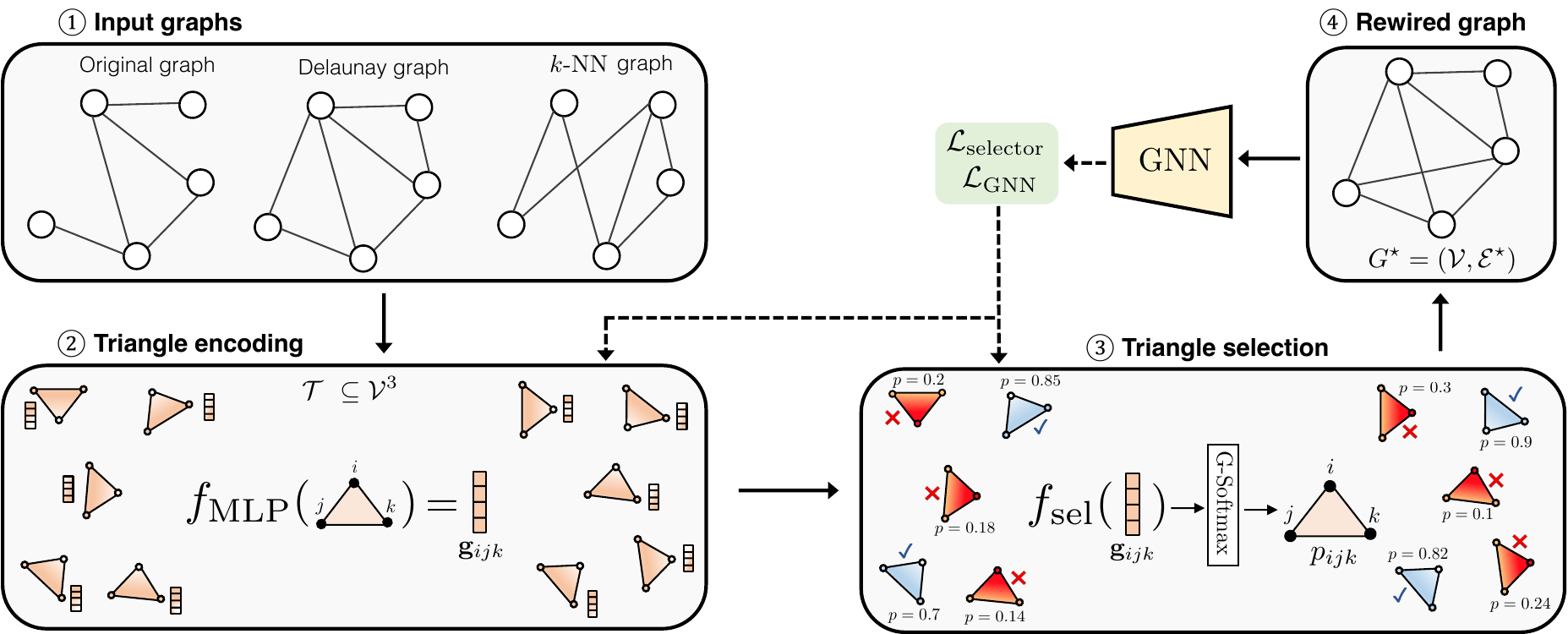}
     \caption{Overview of the TRIGON pipeline. (1) Triangles from the complementary input graphs are aggregated to capture diverse structural signals. (2) A shared encoder maps them into latent representations. (3) A differentiable selector identifies task-relevant triangles to construct a dynamically rewired graph (4) tailored for effective message passing. \label{fig:overview}}
     \label{framework}
 \end{figure*}



\section{Differentiable Graph Rewiring with TRIGON \label{sec:method}}
We introduce TRIGON, a learnable approach to graph rewiring centered on triangle selection. Instead of relying on fixed connectivity rules, it adaptively constructs the graph during training by identifying triangles that support discriminative learning and robust information flow. An overview of TRIGON is given in Fig. \ref{fig:overview}.

\subsection{Triangle Encoding and Selection}
Let \(G = (\mathcal{V}, \mathcal{E})\) be an undirected graph, where each node \(i \in \mathcal{V}\) is equipped with a feature vector \(\mathbf{x}_i \in \mathbb{R}^d\) and a class label \(y_i \in \{0, \dots, C-1\}\). To construct a rich set of candidate triangles, we aggregate triangles from multiple graph views. For sparse real-world graphs, this is often near-linear in practice.
The input graph provides triangles that are explicitly present in the observed topology. A \emph{k}-nearest neighbor ($k$-NN) graph, constructed in the feature space, contributes triangles between similar nodes that may not be connected in the original structure. Finally, applying Delaunay triangulation to the learned embeddings links nodes that are close in the task-informed feature space \cite{attali2024delaunay}. The choice of these three graphs is motivated, on the one hand, by the structural diversity of the induced triangles, each construction capturing complementary information from the original topology and the feature space, and, on the other hand, by their low computational cost, which makes them suitable for integration into a differentiable pipeline.

In what follows, we elaborate on the different components of TRIGON for triangle encoding and selection.

\paragraph{\textbf{(1) Triangle encoding.}}  
Let \( \mathcal{T} \subseteq \mathcal{V}^3 \) be a set of candidate triangles, derived from multiple sources (e.g., the original graph, feature-based $k$-NN graph, and Delaunay triangulation over learned embeddings). For each triangle \( (i, j, k) \in \mathcal{T} \), we construct a raw feature representation via concatenation:
$\mathbf{z}_{ijk} = [\mathbf{x}_i \, \| \, \mathbf{x}_j \, \| \, \mathbf{x}_k] \in \mathbb{R}^{3d}.$
This representation is processed by a triangle encoder network, implemented as a multi-layer perceptron (MLP),
\[
\mathbf{g}_{ijk} = f_{\mathrm{MLP}}(\mathbf{z}_{ijk}) \in \mathbb{R}^{d'},
\]
producing an embedding that captures both geometric and feature-based relationships among the triangle’s vertices.

\paragraph{\textbf{(2) Differentiable triangle selection.}}  
To determine which triangles should contribute to the new graph structure, we pass the embedding \( \mathbf{g}_{ijk} \) through another MLP $f_{\mathrm{sel}}$:
\[
\mathbf{s}_{ijk} = f_{\mathrm{sel}}(\mathbf{g}_{ijk}) \in \mathbb{R}^2,
\]
which outputs logits for a binary selection. We apply the Gumbel-Softmax with temperature \( \tau > 0 \) to obtain a differentiable approximation of the discrete selection: $p_{ijk}$ = \text{GumbelSoftmax}($s_{ijk}, \tau) \in$ [0,1], where \( p_{ijk} \) is the probability that triangle \( (i,j,k) \) is selected in the final rewired graph.
This relaxation facilitates gradient flow through the otherwise discrete selection process by approximating categorical sampling with a continuous, differentiable surrogate \cite{jang2017categorical}. Consequently, the model can effectively learn to prioritize triangles that contribute most to downstream tasks, while remaining fully compatible with gradient-based optimization.

\paragraph{\textbf{(3) Graph reconstruction.}} 
From the set of selected triangles,
\[
\mathcal{T}_{\mathrm{sel}} = \left\{ (i,j,k) \in \mathcal{T} \mid p_{ijk} \geq 0.5 \right\},
\]
we reconstruct the edge set as the union of edges induced by each triangle: $\mathcal{E}^\star = \bigcup_{(i,j,k) \in \mathcal{T}_{\mathrm{sel}}} \{ (i,j), (j,k), (k,i) \}.
$

\noindent The rewired graph \( G^\star = (\mathcal{V}, \mathcal{E}^\star) \) aggregates local and non-local triangles, forming a non-planar structure with enhanced expansion properties. Crucially, \( G^\star \) is dynamically updated during training: at each iteration, a new set of triangles is selected, and corresponding edge set \( \mathcal{E}^\star \) is reconstructed accordingly. The resulting graph serves as input for message passing at each step of the learning process.

\subsection{Loss Functions and Training Pipeline}
To jointly optimize triangle selection and downstream node classification, we define a multi-component loss function designed to align local triangle selection with global structural and task-specific criteria. Let \( \mathcal{T} \) denote the set of candidate triangles and \( \mathcal{T}_{\mathrm{sel}} \subset \mathcal{T} \) the subset selected by the Gumbel--Softmax mechanism. Our training objective integrates four complementary components:

\paragraph{\textbf{(1) Supervised classification loss.}}  
We train a GNN on the reconstructed graph \( G^\star = (\mathcal{V}, \mathcal{E}^\star) \), obtained from the selected triangles. Let \( \hat{y}_i \in \mathbb{R}^C \) denote the predicted class probabilities for node \( i \), and \( y_i \in \{0, \dots, C{-}1\} \) its ground truth label. The node classification loss is the average cross-entropy over the training set:
\[
\mathcal{L}_{\mathrm{GNN}} = \frac{1}{|\mathcal{V}_{\mathrm{train}}|} \sum_{i \in \mathcal{V}_{\mathrm{train}}} \mathrm{CE}(\hat{y}_i, y_i).
\]

\paragraph{\textbf{(2) Contrastive triangle label loss.}}  
To guide the triangle selector \(f_{\mathrm{sel}}\) toward informative class-level motifs, we define a binary supervision signal for each triangle. Specifically, for each \( (i,j,k) \in \mathcal{T} \), we define:
\[
y_{ijk}^{\triangle} = 
\begin{cases}
1 & \text{if at least two nodes among } \{i,j,k\} \text{ share the same label}, \\
0 & \text{otherwise}.
\end{cases}
\]
We then apply a contrastive loss function, encouraging the selection of meaningful triangles:
\[
\mathcal{L}_{\mathrm{contr}} = \frac{1}{|\mathcal{T}|} \sum_{(i,j,k) \in \mathcal{T}} (1 - y_{ijk}^{\triangle}) \cdot p_{ijk}^2 + y_{ijk}^{\triangle} \cdot (0, 1 - p_{ijk}).
\]
The loss aligns triangle selection with supervised signals by prioritizing motifs where labels are partially consistent.

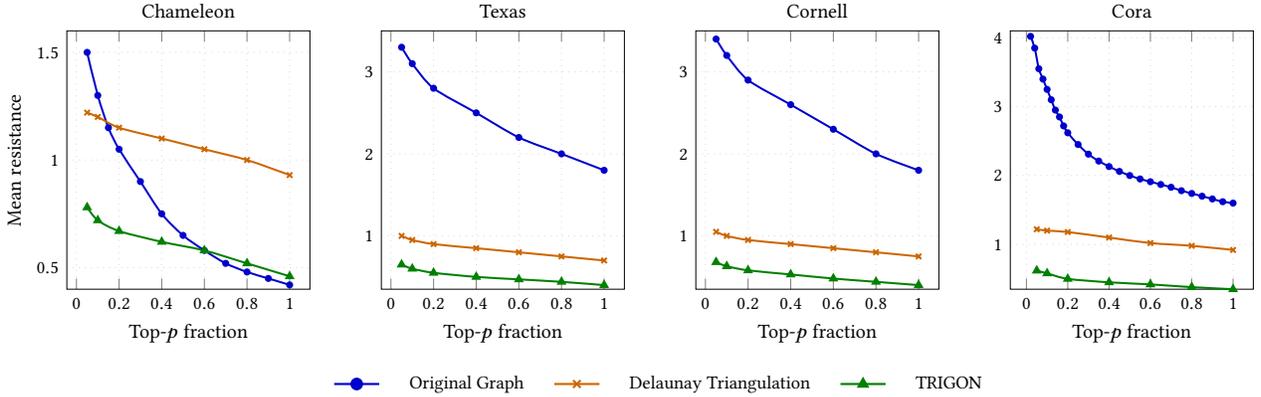
\begin{figure*}[ht]
\centering
\begin{tikzpicture}[scale=0.95]

\begin{groupplot}[
    group style={
        group size=4 by 1,
        horizontal sep=1cm
    },
    width=0.28\textwidth,
    height=5.2cm,
    xlabel={Top-$p$ fraction},
    ylabel={Mean resistance},
    xtick={0,0.2,0.4,0.6,0.8,1.0},
    ytick style={draw=none},
    tick style={line width=0.3pt},
    axis line style={line width=0.3pt},
    tick label style={font=\footnotesize},
    label style={font=\small},
    title style={font=\small, yshift=-1ex},
    grid=major,
    grid style={dotted, gray!30}
]

\nextgroupplot[
    title={Chameleon},
    ymin=0.4, ymax=1.6
]
\addplot[blue!80!black, thick, mark=*, mark size=1pt, smooth] table[row sep=\\] {
0.05 1.5 \\ 0.10 1.3 \\ 0.15 1.15 \\ 0.20 1.05 \\ 0.30 0.9 \\ 0.40 0.75 \\ 0.50 0.65 \\ 0.60 0.58 \\ 0.70 0.52 \\ 0.80 0.48 \\ 0.90 0.45 \\ 1.00 0.42 \\
};
\addplot[orange!80!black, thick, mark=x, mark size=1.5pt, smooth] table[row sep=\\] {
0.05 1.22 \\ 0.10 1.20 \\ 0.20 1.15 \\ 0.40 1.10 \\ 0.60 1.05 \\ 0.80 1.00 \\ 1.00 0.93 \\
};
\addplot[green!50!black, thick, mark=triangle*, mark size=1.5pt, smooth] table[row sep=\\] {
0.05 0.78 \\ 0.10 0.72 \\ 0.20 0.67 \\ 0.40 0.62 \\ 0.60 0.58 \\ 0.80 0.52 \\ 1.00 0.46 \\
};

\nextgroupplot[
    title={Texas},
    ylabel={},
    ymin=0.35, ymax=3.5
]
\addplot[blue!80!black, thick, mark=*, mark size=1pt, smooth] table[row sep=\\] {
0.05 3.3 \\ 0.10 3.1 \\ 0.20 2.8 \\ 0.40 2.5 \\ 0.60 2.2 \\ 0.80 2.0 \\ 1.00 1.8 \\
};
\addplot[orange!80!black, thick, mark=x, mark size=1.5pt, smooth] table[row sep=\\] {
0.05 1.00 \\ 0.10 0.95 \\ 0.20 0.90 \\ 0.40 0.85 \\ 0.60 0.80 \\ 0.80 0.75 \\ 1.00 0.70 \\
};
\addplot[green!50!black, thick, mark=triangle*, mark size=1.5pt, smooth] table[row sep=\\] {
0.05 0.65 \\ 0.10 0.60 \\ 0.20 0.55 \\ 0.40 0.50 \\ 0.60 0.47 \\ 0.80 0.44 \\ 1.00 0.40 \\
};

\nextgroupplot[
    title={Cornell},
    ylabel={},
    ymin=0.35, ymax=3.5
]
\addplot[blue!80!black, thick, mark=*, mark size=1pt, smooth] table[row sep=\\] {
0.05 3.4 \\ 0.10 3.2 \\ 0.20 2.9 \\ 0.40 2.6 \\ 0.60 2.3 \\ 0.80 2.0 \\ 1.00 1.8 \\
};
\addplot[orange!80!black, thick, mark=x, mark size=1.5pt, smooth] table[row sep=\\] {
0.05 1.05 \\ 0.10 1.00 \\ 0.20 0.95 \\ 0.40 0.90 \\ 0.60 0.85 \\ 0.80 0.80 \\ 1.00 0.75 \\
};
\addplot[green!50!black, thick, mark=triangle*, mark size=1.5pt, smooth] table[row sep=\\] {
0.05 0.68 \\ 0.10 0.63 \\ 0.20 0.58 \\ 0.40 0.53 \\ 0.60 0.48 \\ 0.80 0.44 \\ 1.00 0.40 \\
};

\nextgroupplot[
    title={Cora},
    ylabel={},
    ymin=0.35, ymax=4.1
]
\addplot[blue!80!black, thick, mark=*, mark size=1pt, smooth] table[row sep=\\] {
0.02 4.02 \\ 0.04 3.85 \\ 0.06 3.55 \\ 0.08 3.40 \\ 0.10 3.25 \\
0.12 3.10 \\ 0.14 2.95 \\ 0.16 2.85 \\ 0.18 2.72 \\ 0.20 2.62 \\
0.25 2.45 \\ 0.30 2.31 \\ 0.35 2.21 \\ 0.40 2.13 \\ 0.45 2.06 \\
0.50 2.00 \\ 0.55 1.95 \\ 0.60 1.91 \\ 0.65 1.87 \\ 0.70 1.83 \\
0.75 1.78 \\ 0.80 1.74 \\ 0.85 1.70 \\ 0.90 1.66 \\ 0.95 1.62 \\
1.00 1.60 \\
};
\addplot[orange!80!black, thick, mark=x, mark size=1.5pt, smooth] table[row sep=\\] {
0.05 1.22 \\ 0.10 1.20 \\ 0.20 1.18 \\ 0.40 1.10 \\
0.60 1.02 \\ 0.80 0.98 \\ 1.00 0.92 \\
};
\addplot[green!50!black, thick, mark=triangle*, mark size=1.5pt, smooth] table[row sep=\\] {
0.05 0.62 \\ 0.10 0.58 \\ 0.20 0.50 \\ 0.40 0.45 \\
0.60 0.42 \\ 0.80 0.38 \\ 1.00 0.35 \\
};

\end{groupplot}

\path (group c1r1.south west) -- node[below=0.9cm, draw=none] {
    \begin{minipage}{0.8\textwidth}
    \centering
    \begin{tikzpicture}
    \begin{axis}[
        hide axis,
        xmin=0, xmax=1,
        ymin=0, ymax=1,
        legend columns=3,
        legend style={draw=none, font=\footnotesize, column sep=1em}
    ]
    \addlegendimage{blue!80!black, thick, mark=*}
    \addlegendentry{Original Graph}
    \addlegendimage{orange!80!black, thick, mark=x}
    \addlegendentry{Delaunay Triangulation}
    \addlegendimage{green!50!black, thick, mark=triangle*}
    \addlegendentry{TRIGON}
    \end{axis}
    \end{tikzpicture}
    \end{minipage}
} (group c4r1.south east);

\end{tikzpicture}
\caption{Mean resistance among top high-resistance edges across four datasets. TRIGON achieves the lowest resistance in critical edges, indicating improved global connectivity.}
\label{fig:mean_resistance}
\end{figure*}

\paragraph{\textbf{(3) Structural smoothness loss.}}  
We introduce a structural regularizer that encourages the selection of geometrically coherent triangles. For each selected triangle \( (i,j,k) \in \mathcal{T}_{\mathrm{sel}} \), we compute:
\[
\mathcal{L}_{\mathrm{struct}} = \frac{1}{|\mathcal{T}_{\mathrm{sel}}|} \sum_{(i,j,k) \in \mathcal{T}_{\mathrm{sel}}} \left( \|\mathbf{x_i} - \mathbf{x_j}\|_2 + \|\mathbf{x_j} - \mathbf{x_k}\|_2 + \|\mathbf{x_k} - \mathbf{x_i}\|_2 \right).
\]
This term acts as a soft constraint to favor triangles composed of semantically or geometrically close nodes in feature space.

\paragraph{\textbf{(4) Class-wise participation regularization.}}  
Let \( \mathcal{T}_i \subseteq \mathcal{T}_{\text{sel}} \) denote the set of selected triangles in which node \( i \) participates. We define a regularization term that enforces consistency in triangle participation across nodes sharing the same class. Thus, we associate each class \( c\) with a learnable scalar \( \pi_c \in \mathbb{R} \), representing the target number of triangles for nodes of class \( c \). The loss is given by:
\[
\mathcal{L}_{\mathrm{part}} = \frac{1}{|\mathcal{V}_{\mathrm{train}}|} \sum_{i \in \mathcal{V}_{\mathrm{train}}} \left( |\mathcal{T}_i| - \pi_{y_i} \right)^2.
\]
This objective serves two complementary purposes. First, it mitigates structural imbalance in the induced topology. Without such regulation, nodes from certain classes may be underrepresented in the rewired graph. By encouraging a more uniform participation of nodes across all classes, the model promotes a structurally balanced graph topology that better supports class discrimination. Second, in heterophilic settings, enabling class-specific participation  allows the model to learn differentiated roles for each class. This flexibility is especially beneficial when classes require different topological configurations to effectively propagate information \cite{wang2024understanding}.

\paragraph{\textbf{(5) Total loss and joint optimization.}}  

The training proceeds in two intertwined stages at each epoch. First, we optimize triangle selection by minimizing a composite loss:
\begin{equation}
\mathcal{L}_{\text{selector}} = \mathcal{L}_{\text{contr}} + \mathcal{L}_{\text{part}} + \mathcal{L}_{\text{struct}}.
\label{eq:total_loss}
\end{equation}
This updates the triangle encoder $f_{\text{MLP}}$ and selector $f_{\text{sel}}$, using the current GNN embeddings. The selected triangles induce a new graph $G^\star$, which is then used to train the GNN with a standard supervised objective. Each epoch alternates between these two steps, gradually refining both the topology and the node representations.

\paragraph{\textbf{Computational complexity.}}  
TRIGON adds minimal overhead to the backbone architecture, as it preserves the original GNN operations and introduces additional computations only for triangle extraction and selection. Triangle enumeration is performed efficiently via neighbor intersection in \(\mathcal{O}(a \cdot m)\) time, where \(a\) is the graph's arboricity~\cite{chiba1985arboricity}, while the selection module relies on lightweight, parallelizable MLPs. As a result, the overall scalability of the method remains comparable to that of standard GNNs.

\begin{table}[t]
\centering
\begin{tabular}{rcccc}
\toprule
\textbf{Dataset} & \textbf{DR diam.} & \textbf{TRIGON diam.} & $\lambda_2$ \textbf{improv.} \\
\midrule
Texas     & 11 & 7 & $303\%$ \\
Cornell   & 12 & 6 &$424\%$ \\
Wisconsin & 12 & 8 &$236\%$ \\
Chameleon & 21 & 10 &$ 1,220\%$ \\
Actor     & 49 & 12 & $ 18,966\%$ \\
Squirrel    & 34 & 10 & $ 4,081\%$ \\
Roman-Empire     & 65 & 25 &$ 1,550\%$ \\
\bottomrule
\end{tabular}
\caption{Comparison of Delaunay triangulation rewiring (DR) diameter and spectral gap $\lambda_2$ improvement. TRIGON achieves a lower diameter and a higher $\lambda_2$ than DR on all datasets.}

\label{tab:diam_gap}
\end{table}

\subsection{Structural Properties of TRIGON-Rewired Graphs}

Before presenting the experimental results in Section \ref{sec:experiments}, we compare the structural properties of the triangulation produced by our framework with those of the classical Delaunay triangulation. This includes the examination of spectral properties and the effective resistance, with the aim of understanding the effect of TRIGON on the structure of the rewired graphs.

To assess the presence and severity of structural bottlenecks in different graph constructions, we analyze the effective resistance across edges. Specifically, we plot the mean effective resistance among the top-$p$ fraction of edges with the highest resistance values, for varying values of $p \in [0,1]$. This diagnostic focuses on the worst-case communication paths, where high resistance indicates restricted flow of information, typically edges which accentuate oversquashing \cite{resit}.  As shown in Fig.~\ref{fig:mean_resistance}, the TRIGON-rewired graph consistently exhibits lower mean effective resistance in the highest-resistance quantiles compared to both the original graph and the Delaunay rewiring. This reduction is particularly evident for low values of $p$, which correspond to the most critical edges, suggesting that TRIGON helps to mitigate such structural bottlenecks.

\begin{table*}[ht!]
    \begin{center}
        \small
        \begin{tabular}{rccccccccccc}
            \toprule
            \textbf{} & \textbf{Base (GCN)} & \textbf{DIGL} & \textbf{FA} & \textbf{SRDF} & \textbf{FOSR} & \textbf{BORF} & \textbf{GTR}  & \textbf{JDR} & \textbf{DR} & \textbf{TRIGON} \\
            \midrule
            Cham. & 65.35$\pm$0.54 & 54.82$\pm$0.48 & 26.34$\pm$0.61 & 63.08$\pm$0.37 & 67.98$\pm$0.40 & 65.35$\pm$0.51 & 68.03$\pm$0.61 & 65.85$\pm$0.49 & \underline{74.28}$\pm$0.48 & \textbf{75.52}$\pm$0.50 \\
            Squir. & 51.30$\pm$0.38 & 40.53$\pm$0.29 & 22.88$\pm$0.42 & 49.11$\pm$0.28 & 52.63$\pm$0.30 & $\ge$24h & 53.32$\pm$0.44 & 53.78  $\pm$0.46 & \underline{65.25}$\pm$0.26 & \textbf{66.48}$\pm$0.35\\
            Actor & 30.02$\pm$0.22 & 26.75$\pm$0.23 & 26.03$\pm$0.30 & {31.85}$\pm$0.22 & 29.26$\pm$0.23 & 31.36$\pm$0.27 & 31.08$\pm$0.28 & 34.12 $\pm$0.33 & \underline{41.36}$\pm$0.20 & \textbf{43.81}$\pm$0.24 \\
            Texas & 56.19$\pm$1.61 & 45.95$\pm$1.58 & 55.93$\pm$1.76 & 59.79$\pm$1.71 & 61.35$\pm$1.25 & 56.30$\pm$1.61 & 57.18$\pm$1.64 & 69.56 $\pm$1.71 & \underline{70.46}$\pm$1.61 & \textbf{75.74}$\pm$1.61\\
            Wisc. & 55.12$\pm$1.51 & 46.90$\pm$1.28 & 46.77$\pm$1.48 & 58.49$\pm$1.23 & 55.60$\pm$1.25 & 55.37$\pm$1.47 & 57.22$\pm$1.50 & 67.87 $\pm$1.62 & \underline{70.98}$\pm$1.50 & \textbf{73.90}$\pm$1.61\\
            Cornell & 44.78$\pm$1.45 & 44.46$\pm$1.37 & 45.33$\pm$1.55 & {47.73}$\pm$1.51 & 45.11$\pm$1.47 & 46.81$\pm$1.56 & 47.57$\pm$1.52 & 57.31 $\pm$1.60 & \underline{67.22}$\pm$1.48 & \textbf{69.11}$\pm$1.53 \\
            R-Emp. & 51.66$\pm$0.17 & 53.93$\pm$0.14 & OOM & 52.53$\pm$0.13 & 52.38$\pm$0.21 & {58.58}$\pm$0.14 & 53.31$\pm$0.23 & \textbf{71.23} $\pm$0.18 & {61.99}$\pm$0.14 &  \underline{66.52}$\pm$0.13 \\
            \midrule
            Cora & 87.73$\pm$0.25 & \underline{88.31}$\pm$0.29 & 29.86$\pm$0.28 & 87.73$\pm$0.31 & 87.94$\pm$0.26 & 87.72$\pm$0.27 & 87.86$\pm$0.28 & 87.54$\pm$0.25 & \underline{91.39}$\pm$0.24 & \textbf{91.71}$\pm$0.29 \\
            Citeseer & 76.01$\pm$0.25 & 76.22$\pm$0.34 & 22.31$\pm$0.34 & 76.43$\pm$0.32 & 76.34$\pm$0.27 & \underline{76.49}$\pm$0.28 & 76.12$\pm$0.28 & 76.09 $\pm$0.29 & \underline{81.14}$\pm$0.34 & \textbf{82.85}$\pm$0.38\\
            Pubmed & 88.20$\pm$0.10 & \underline{88.51}$\pm$0.10 & OOM & 88.16$\pm$0.11 & 88.42$\pm$0.10 & 88.34$\pm$0.10 & 88.44$\pm$0.10 & 88.14$\pm$0.10 & \underline{88.69}$\pm$0.10 & \textbf{90.01}$\pm$0.13 \\
            \bottomrule
        \end{tabular}
        \caption{Experimental results (accuracy) on \textbf{heterophilic} and \textbf{homophilic} datasets with \textbf{GCN} as backbone. Best score in bold and second-best score underlined.}
        \label{HeteroGCN}
    \end{center}
\end{table*}

\begin{table*}[ht!]
    \begin{center}
        \small
        \begin{tabular}{rccccccccccc}
            \toprule
            \textbf{{}} & \textbf{Base (GAT)} & \textbf{DIGL} & \textbf{FA} & \textbf{SRDF} & \textbf{FOSR} & \textbf{BORF} & \textbf{GTR} & \textbf{JDR} & \textbf{DR} & \textbf{TRIGON} \\
            \midrule
            Cham. & 65.07$\pm$0.41 & 56.34$\pm$0.43 & 27.11$\pm$0.56 & 63.15$\pm$0.44 & 66.61$\pm$0.45 & 66.92$\pm$0.51 & 65.97$\pm$0.54 & 65.30 $\pm$0.59 & \underline{72.04}$\pm$0.37 & \textbf{75.54}$\pm$0.58 \\
            Squir. & 50.87$\pm$0.56 & 41.65$\pm$0.68 & 21.49$\pm$0.71 & 50.36$\pm$0.38 & 52.02$\pm$0.43 & $\geq 24h$ & 52.72$\pm$0.48 & 51.21$\pm$0.64 & \underline{61.47}$\pm$0.29 & \textbf{66.12}$\pm$0.40 \\
            Actor & 29.92$\pm$0.23 & 31.22$\pm$0.47 & 28.20$\pm$0.51 & 31.47$\pm$0.25 & 29.73$\pm$0.24 & 29.64$\pm$0.33 & 30.13$\pm$0.31 & 32.71 $\pm$0.40 & \underline{40.25}$\pm$0.23 & \textbf{44.02}$\pm$0.28 \\
            Texas & 56.84$\pm$1.61 & 46.49$\pm$1.63 & 56.17$\pm$1.71 & 57.45$\pm$1.62 & 61.85$\pm$1.41 & 56.68$\pm$1.49 & 57.88$\pm$1.65 & 64.75 $\pm$1.65 & \underline{74.30}$\pm$1.38 & \textbf{77.29}$\pm$1.55 \\
            Wisc. & 53.58$\pm$1.39 & 46.29$\pm$1.47 & 46.95$\pm$1.52 & 56.80$\pm$1.29 & 54.06$\pm$1.27 & 55.39$\pm$1.23 & 56.53$\pm$1.64 & 60.06 $\pm$1.45 & \underline{74.33}$\pm$1.24 & \textbf{75.81}$\pm$1.38 \\
            Cornell & 46.05$\pm$1.49 & 44.05$\pm$1.44 & 44.60$\pm$1.74 & 48.03$\pm$1.66 & 48.30$\pm$1.61 & 48.57$\pm$1.56 & 48.70$\pm$1.63 & 58.19$\pm$1.58& \underline{68.03}$\pm$1.62 & \textbf{70.63}$\pm$1.66 \\
            R-Emp. & 49.23$\pm$0.33 & 53.89$\pm$0.16 & OOM & 50.75$\pm$0.17 & 49.54$\pm$0.31 & 51.03$\pm$0.26 & 50.60$\pm$0.24 & \underline{62.09} $\pm$0.18  & 61.80$\pm$0.16 & \textbf{64.36}$\pm$0.20 \\
            \midrule
            Cora & 87.65$\pm$0.24 & 88.31$\pm$0.29 & 30.44$\pm$0.26 & 88.11$\pm$0.28 & 88.13$\pm$0.27 & 87.72$\pm$0.27 & 87.94$\pm$0.23 & 87.91 $\pm$0.25 & \underline{91.37}$\pm$0.23 & \textbf{91.71}$\pm$0.29 \\
            Citeseer & 76.20$\pm$0.27 & 76.22$\pm$0.34 & 23.11$\pm$0.32 & 76.26$\pm$0.31 & 75.94$\pm$0.32 & 76.44$\pm$0.44 & 76.35$\pm$0.28 & 77.80 $\pm$0.39 & \underline{81.61}$\pm$0.25 & \textbf{82.85}$\pm$0.38 \\
            Pubmed & 87.39$\pm$0.11 & 87.96$\pm$0.10 & OOM & 87.44$\pm$0.12 & 87.56$\pm$0.11 & 87.61$\pm$0.12 & 87.31$\pm$0.12 & 87.73 $\pm$0.10 & \underline{89.14}$\pm$0.09 & \textbf{90.01}$\pm$0.13 \\
            \bottomrule
        \end{tabular}
        \caption{Experimental results (accuracy) on \textbf{heterophilic} and \textbf{homophilic} datasets with \textbf{GAT} as backbone. Best score in bold and second-best score underlined.}
        \label{HeteroGAT}
    \end{center}
\end{table*}
The results validate that augmenting the graph with carefully selected 
non-local triangles leads to improved global connectivity and more effective message propagation in GNNs.
As we will observe in Section~\ref{sec:experiments}, this in turn leads to more effective message propagation and better downstream performance in GNNs.

We have further examined the diameter and spectral gap of the rewired graphs. As shown in Table~\ref{tab:diam_gap}, our TRIGON framework achieves a consistent reduction in graph diameter compared to the Delaunay graph. This effect becomes increasingly pronounced on larger graphs, where long-range dependencies are critical for effective learning. By selecting both local and distant triangles, TRIGON overcomes the inherent planarity and locality constraints of Delaunay constructions, introducing structurally meaningful shortcuts that shrink path lengths between distant nodes. These structural modifications lead to enhanced expansion capabilities, reflected by the observed increase in the spectral gap. Specifically, according to Eq.~\eqref{diam}, the Cheeger constant provides a lower bound on the spectral gap in terms of the graph's diameter and minimum degree. Thus, the observed diameter reduction directly implies stronger spectral connectivity, enabling more effective propagation of information across the graph and alleviating the limitations of traditional message passing on sparse or high-diameter topologies.

\section{Experimental Evaluation} \label{sec:experiments}

We have conducted experiments on ten different datasets for the node classification task, comprising seven heterophilic datasets \cite{tang2009social, rozemberczki2021multi, platonov2023critical} and three homophilic datasets \cite{DATASET_SCIENT}.

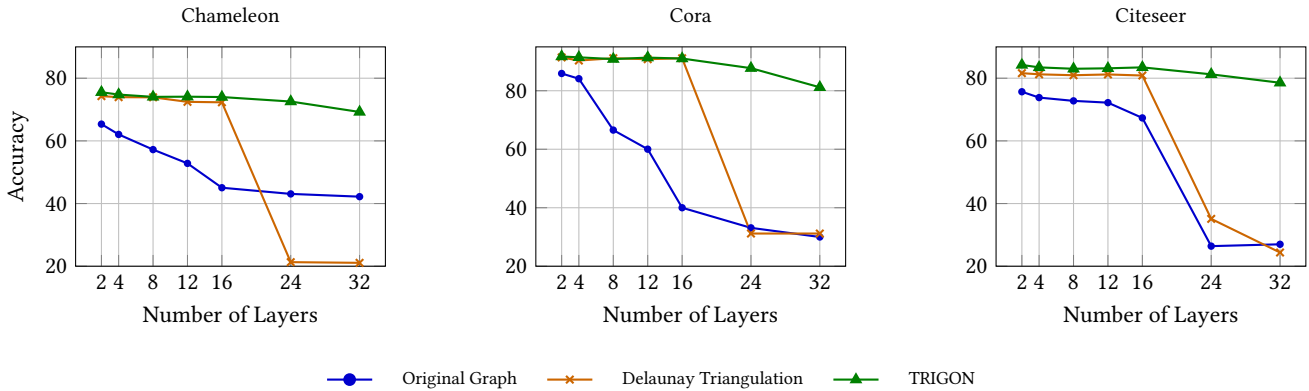
\begin{figure*}[!ht]
\centering
\begin{tikzpicture}
\begin{groupplot}[
  group style={
    group size=3 by 1,
    horizontal sep=2cm
  },
  width=5.7cm,
  height=4.5cm,
  grid=major,
  axis line style={thin},
  xlabel={Number of Layers},
  xtick={2,4,8,12,16,24,32},
  ymin=25, ymax=95
]

\nextgroupplot[
  title={\small Chameleon},
  ylabel={Accuracy},
  ymin=20, ymax=90
]
\addplot[blue!80!black, thick, mark=*,mark size=1pt] coordinates {
  (2, 65.35) (4, 62.06) (8, 57.22) (12, 52.82) (16, 45.05) (24, 43.08) (32, 42.20)
};
\addplot[orange!80!black, thick, mark=x] coordinates {
  (2, 74.28) (4, 73.95) (8, 73.90) (12, 72.46) (16, 72.31) (24, 21.30) (32, 21.08)
};
\addplot[green!50!black, thick, mark=triangle*] coordinates {
  (2, 75.52) (4, 74.78) (8, 74.05) (12, 74.10) (16, 73.99) (24, 72.56) (32, 69.23)
};

\nextgroupplot[
  title={\small Cora},
  ymin=20, ymax=95
]
\addplot[blue!80!black, thick, mark=*,mark size=1pt] coordinates {
  (2, 85.90) (4, 84.11) (8, 66.55) (12, 60.02) (16, 40.00) (24, 33.12) (32, 30.00)
};
\addplot[orange!80!black, thick, mark=x] coordinates {
  (2, 91.39) (4, 90.33) (8, 91.12) (12, 90.78) (16, 91.11) (24, 31.18) (32, 31.18)
};
\addplot[green!50!black, thick, mark=triangle*] coordinates {
  (2, 91.71) (4, 91.43) (8, 90.82) (12, 91.33) (16, 91.01) (24, 87.72) (32, 81.18)
};

\nextgroupplot[
  title={\small Citeseer},
  ymin=20, ymax=90
]
\addplot[blue!80!black, thick, mark=*,mark size=1pt] coordinates {
  (2, 75.68) (4, 73.82) (8, 72.77) (12, 72.20) (16, 67.33) (24, 26.41) (32, 27.00)
};
\addplot[orange!80!black, thick, mark=x] coordinates {
  (2, 81.61) (4, 81.24) (8, 80.95) (12, 81.20) (16, 80.86) (24, 35.11) (32, 24.38)
};
\addplot[green!50!black, thick, mark=triangle*] coordinates {
  (2, 84.23) (4, 83.46) (8, 83.01) (12, 83.16) (16, 83.46) (24, 81.24) (32, 78.55)
};

\end{groupplot}
\end{tikzpicture}

\vspace{3mm}

\begin{tikzpicture}
\begin{axis}[
    hide axis,
    xmin=0, xmax=1,
    ymin=0, ymax=1,
    legend columns=3,
    legend style={draw=none, font=\footnotesize, column sep=1em}
]
\addlegendimage{blue!80!black, thick, mark=*}
\addlegendentry{Original Graph}
\addlegendimage{orange!80!black, thick, mark=x}
\addlegendentry{Delaunay Triangulation}
\addlegendimage{green!50!black, thick, mark=triangle*}
\addlegendentry{TRIGON}
\end{axis}
\end{tikzpicture}

\caption{Effect of GCN model depth on classification accuracy across Chameleon, Cora, and Citeseer. TRIGON outperforms both the original and Delaunay-rewired graphs, showing greater robustness to oversmoothing}
\label{oversm}
\end{figure*}

\subsection{Baseline Models}

We compare TRIGON against eight rewiring methods designed to mitigate oversquashing or enhance graph connectivity. FA \cite{alon2020bottleneck} introduces fully connected connections at the final GNN layer, aiming to alleviate oversquashing by enabling global communication. DIGL \cite{DIGL}\footnote{https://github.com/gasteigerjo/gdc} enhances connectivity through diffusion-based edge augmentation inspired by personalized PageRank. SDRF \cite{UNDERSTANDING_bottlenecks}\footnote{https://github.com/jctops/understanding-over-squashing/tree/main} builds upon Ricci curvature, proposing a stochastic discrete Ricci Flow that rewires the graph by balancing negatively curved edges. FOSR \cite{FOSR}\footnote{https://github.com/kedar2/FoSR/tree/main} selects edges that maximize a first-order approximation of the spectral gap, promoting more efficient message propagation. BORF \cite{BORF}\footnote{https://github.com/hieubkvn123/revisiting-gnn-curvature} tackles both oversmoothing and oversquashing by pruning edges with extreme curvature values. GTR \cite{resit}\footnote{https://github.com/blackmit/gtr\_rewiring} is
an iterative algorithm to add edges to the graph to minimize total resistance. DR \cite{attali2024delaunay}\footnote{https://github.com/Hugo-Attali/Delaunay-Rewiring} leverages node features to perform Delaunay triangulation-based rewiring. Finally, JDR\cite{linkerhagner2025joint}\footnote{https://github.com/jlinki/JDR/tree/main} reconstructs the graph to maximize spectral alignment between structural and feature information.

\subsection{Experimental Setup}
To optimize the triangle selector module $f_{\text{sel}}$, we perform a grid search over the following hyperparameters: learning rate between $\{0.01, 0.005, 0.001\}$, weight decay values between $\{5\text{e}^{-5}, 5\text{e}^{-6}\}$, and hidden dimensions between $\{64, 128, 256\}$. For the construction of the $k$-nearest neighbor ($k$-NN) graph, we set the number of neighbors to \( k \in \{10, 20\} \), depending on the dataset size and the density of the feature space.
For the GNN backbones, we use two standard GNN architectures, GCN \cite{gcn} and GAT \cite{gat}, to compare several graph rewiring strategies. We adopt the training framework introduced in \cite{pei2020geom,attali2024delaunay}.  Specifically, we set the number of layers to $2$, dropout rate to $0.5$, learning rate to $0.005$, patience to $100$ epochs, and weight decay to $5\text{e}^{-6}$ for Texas, Wisconsin, and Cornell, or $5\text{e}^{-5}$ for all other datasets. The hidden dimension is set to $32$ for Texas, Wisconsin, Cornell, and Actor; $48$ for Squirrel, Chameleon, and Roman-Empire; and $16$ for Cora, Citeseer, and Pubmed. For all datasets, we follow the same data split: $60\%$ of nodes are used for training, $20\%$ for validation, and the remaining $20\%$ for testing.

\subsection{Results}
The node classification results are shown in Tables \ref{HeteroGCN} and \ref{HeteroGAT}. As we can observe, TRIGON  outperforms state-of-the-art graph rewiring techniques across nine out of ten evaluated benchmarks, regardless of the backbone (GCN or GAT) and under both homophilic and heterophilic conditions. On average, applying a standard GCN on TRIGON-rewired graphs yields a classification accuracy improvement exceeding \textbf{25\%} compared to the original graph.

More broadly, these results confirm that feature-based rewiring can provide a significant advantage over purely structural methods, as also observed in \cite{attali2024delaunay}, where feature-aware approaches such as TRIGON and DR consistently outperform structural rewiring baselines. In particular, TRIGON goes beyond static approaches like Delaunay-based rewiring (DR) by learning to select non-local triangles that better align with task-specific dependencies. This adaptivity enables the construction of structurally rich triangulations, improving spectral expansion and facilitating more effective long-range message propagation.

\subsection{Oversmoothing Analysis}
Oversmoothing, typically linked to deeper GNNs~\cite{oversmoothing,Over-Smoothing_gnn}, can also arise when structural neighborhoods closely align with feature similarity. This is notably the case in Delaunay-rewired graphs, where triangles are formed between spatially adjacent nodes, often reinforcing local redundancy. Such configurations may accelerate the convergence of node embeddings toward indistinguishable node representations, sometimes after only a few layers~\cite{chen2020measuring}. Since TRIGON dynamically selects triangles based on their contribution to the learning objective, it induces more diverse and task-relevant connectivity patterns. In particular, the inclusion of both local and long-range triangles diversifies message passing and helps preserve feature variability. We therefore investigate whether our method offers greater robustness to oversmoothing.

Figure \ref{oversm} reports classification accuracy versus GCN depth on Chameleon, as well as on the homophilic datasets Cora and Citeseer, where oversmoothing is known to be more pronounced \cite{zhao2019pairnorm}. As expected, a standard GCN exhibits a steep decline in performance as depth increases, confirming the presence of oversmoothing. Delaunay-based rewiring (DR) provides improvement but still suffers from degradation beyond a few layers. In contrast, TRIGON consistently yields higher accuracy across all depths, indicating a stronger preservation of representational diversity. These findings suggest that task-aware, non-local triangle selection can effectively delay or mitigate oversmoothing, while also improving global connectivity and information propagation.

\subsection{Ablation Studies}

We conduct ablation experiments to evaluate the contribution of the triangle selection mechanism and the associated loss components of TRIGON. In addition, we assess the impact of the candidate triangle sources, the original graph and the $k$-NN graph, to quantify the benefit of integrating structural and feature-based information.

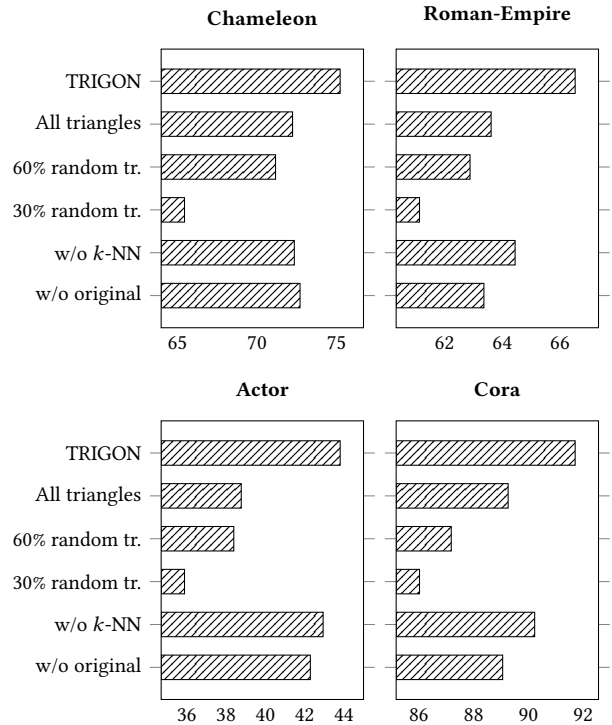
\begin{figure}[t]
  \centering

  \begin{minipage}[b]{0.48\linewidth}
    \centering
    \begin{tikzpicture}
      \begin{axis}[
        width=1.05\linewidth,
        height=1.3\linewidth,
        bar width=0.08\linewidth,
        xbar,
        enlargelimits=0.15,
        yticklabels={{
          w/o original},{w/o $k$-NN},{30\% random tr.},
          {60\% random tr.},{All triangles},{TRIGON}
        },
        ytick=data,
        title={\textbf{Chameleon}},
        tick label style={font=\small},
        label style={font=\scriptsize},
        title style={font=\small},
        xtick style={draw=none}
      ]
        \addplot[pattern=north east lines] coordinates {
          (72.70,0) (72.34,1) (65.44,2)
          (71.16,3) (72.23,4) (75.22,5)
        };
      \end{axis}
    \end{tikzpicture}
  \end{minipage}
  \hfill
  \begin{minipage}[b]{0.48\linewidth}
    \centering
    \begin{tikzpicture}
      \begin{axis}[
        width=1.05\linewidth,
        height=1.3\linewidth,
        bar width=0.08\linewidth,
        xbar,
        enlargelimits=0.15,
        yticklabels={,,,,,},
        ytick=data,
        title={\textbf{Roman-Empire}},
        tick label style={font=\small},
        label style={font=\scriptsize},
        title style={font=\small},
        xtick style={draw=none}
      ]
        \addplot[pattern=north east lines] coordinates {
          (63.36,0) (64.44,1) (61.13,2)
          (62.88,3) (63.61,4) (66.52,5)
        };
      \end{axis}
    \end{tikzpicture}
  \end{minipage}

  \vspace{0.6em}

  \begin{minipage}[b]{0.48\linewidth}
    \centering
    \begin{tikzpicture}
      \begin{axis}[
        width=1.05\linewidth,
        height=1.3\linewidth,
        bar width=0.08\linewidth,
        xbar,
        enlargelimits=0.15,
        yticklabels={{
          w/o original},{w/o $k$-NN},{30\% random tr.},
          {60\% random tr.},{All triangles},{TRIGON}
        },
        ytick=data,
        title={\textbf{Actor}},
        tick label style={font=\small},
        label style={font=\scriptsize},
        title style={font=\small},
        xtick style={draw=none}
      ]
        \addplot[pattern=north east lines] coordinates {
          (42.29,0) (42.94,1) (35.88,2)
          (38.39,3) (38.77,4) (43.81,5)
        };
      \end{axis}
    \end{tikzpicture}
  \end{minipage}
  \hfill
  \begin{minipage}[b]{0.48\linewidth}
    \centering
    \begin{tikzpicture}
      \begin{axis}[
        width=1.05\linewidth,
        height=1.3\linewidth,
        bar width=0.08\linewidth,
        xbar,
        enlargelimits=0.15,
        yticklabels={,,,,,},
        ytick=data,
        title={\textbf{Cora}},
        tick label style={font=\small},
        label style={font=\scriptsize},
        title style={font=\small},
        xtick style={draw=none}
      ]
        \addplot[pattern=north east lines] coordinates {
          (89.06,0) (90.23,1) (86.02,2)
          (87.18,3) (89.26,4) (91.71,5)
        };
      \end{axis}
    \end{tikzpicture}
  \end{minipage}

  \caption{Impact of different triangle selection strategies on node classification accuracy. TRIGON  outperforms both random and full rewiring (All triangles) baselines across datasets. \label{fig:ablation_triangles}}
\end{figure}

\paragraph{\textbf{{Triangle selection}}} We conduct an ablation study to assess the importance of the learned triangle selection mechanism. Specifically, we compare our full method against three variants: (i) random selection of 30\% of the candidate triangles, (ii) random selection of 60\%, and (iii) the inclusion of all candidate triangles without selection (all triangles). Additionally, we evaluate the impact of excluding triangles from either the $k$-NN graph or the original graph structure. The results are presented in Fig.~\ref{fig:ablation_triangles}. We observe a progressive improvement in performance as more triangles are added, indicating that increased connectivity generally benefits the classification task. Notably, our proposed selection strategy consistently achieves the best results, demonstrating that performance gains stem not simply from the number of triangles but from the relevance and effectiveness of those selected.
Furthermore, ablating triangles from either the original graph or the $k$-NN graph leads to performance degradation across all datasets. This highlights the importance of preserving both the structural priors of the original topology and the feature-based proximity relations, which together provide complementary information essential for effective graph rewiring.

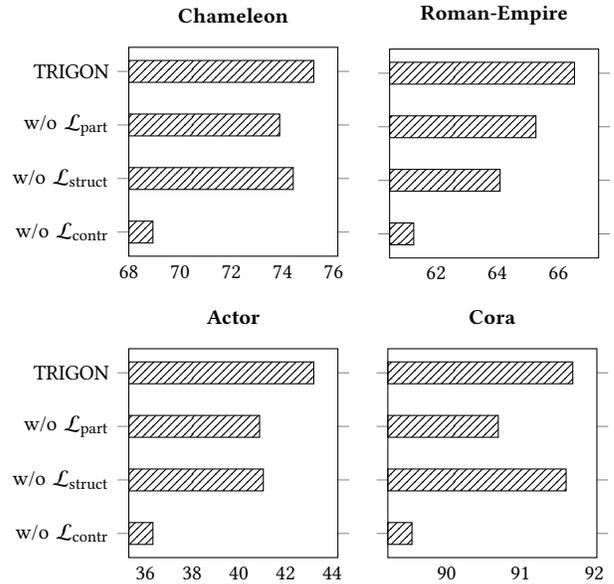
\begin{figure}[h!]
\centering

\begin{minipage}{0.49\linewidth}
\centering
\begin{tikzpicture}
\begin{axis}[
    width=1.05\linewidth,
    height=1.05\linewidth,
    bar width=0.07\linewidth,
    xbar,
    enlargelimits=0.15,
    yticklabels={
        {w/o $\mathcal{L}_{\mathrm{contr}}$},
        {w/o $\mathcal{L}_{\mathrm{struct}}$},
        {w/o $\mathcal{L}_{\mathrm{part}}$},
        {TRIGON}
    },
    ytick=data,
    title={\textbf{Chameleon}},
    tick label style={font=\small},
    label style={font=\scriptsize},
    title style={font=\small},
    xtick style={draw=none}
]
\addplot[pattern=north east lines] coordinates {
    (68.94, 0) (74.42, 1) (73.89, 2) (75.22, 3)
};
\end{axis}
\end{tikzpicture}
\end{minipage}
\hfill
\begin{minipage}{0.49\linewidth}
\centering
\begin{tikzpicture}
\begin{axis}[
    width=1.05\linewidth,
    height=1.05\linewidth,
    bar width=0.07\linewidth,
    xbar,
    enlargelimits=0.15,
    yticklabels={
    },
    ytick=data,
    title={\textbf{Roman-Empire}},
    tick label style={font=\small},
    label style={font=\scriptsize},
    title style={font=\small},
    xtick style={draw=none}
]
\addplot[pattern=north east lines] coordinates {
    (61.27, 0) (64.09, 1) (65.26, 2) (66.52, 3)
};
\end{axis}
\end{tikzpicture}
\end{minipage}

\vspace{0.5em}

\begin{minipage}{0.49\linewidth}
\centering
\begin{tikzpicture}
\begin{axis}[
    width=1.05\linewidth,
    height=1.05\linewidth,
    bar width=0.07\linewidth,
    xbar,
    enlargelimits=0.15,
    yticklabels={
        {w/o $\mathcal{L}_{\mathrm{contr}}$},
        {w/o $\mathcal{L}_{\mathrm{struct}}$},
        {w/o $\mathcal{L}_{\mathrm{part}}$},
        {TRIGON}
    },
    ytick=data,
    title={\textbf{Actor}},
    tick label style={font=\small},
    label style={font=\scriptsize},
    title style={font=\small},
    xtick style={draw=none}
]
\addplot[pattern=north east lines] coordinates {
    (36.32, 0) (41.05, 1) (40.89, 2) (43.21, 3)
}; 
\end{axis}
\end{tikzpicture}
\end{minipage}
\hfill
\begin{minipage}{0.49\linewidth}
\centering
\begin{tikzpicture}
\begin{axis}[
    width=1.05\linewidth,
    height=1.05\linewidth,
    bar width=0.07\linewidth,
    xbar,
    enlargelimits=0.15,
    yticklabels={
    },
    ytick=data,
    title={\textbf{Cora}},
    tick label style={font=\small},
    label style={font=\scriptsize},
    title style={font=\small},
    xtick style={draw=none}
]
\addplot[pattern=north east lines] coordinates {
    (89.53, 0) (91.62, 1) (90.70, 2) (91.71, 3)
};
\end{axis}
\end{tikzpicture}
\end{minipage}

\caption{Impact of each loss component on classification accuracy across four datasets.}
\label{ablation_loss}
\end{figure}

\paragraph{\textbf{Loss components.}} 
Figure~\ref{ablation_loss} reports the effect of removing each of the loss terms from the training objective of Eq. \eqref{eq:total_loss}. The results demonstrate that all components of the loss function are necessary to guide the selection toward structurally coherent and task-relevant rewiring. This suggests that supervision plays an important role in shaping a rewired topology that supports effective message passing.







\section{Conclusion}
In this work, we introduced TRIGON, a framework for graph rewiring via dynamic selection of triangles to address oversquashing and oversmoothing in GNNs. Leveraging a differentiable triangle-based rewiring, TRIGON captures information from both features and graph structure, and learns to construct non-planar triangulations that enhance global connectivity.
We show that theoretical analyses motivate this approach, highlighting the relevance of triangle-based rewiring for improving message propagation through better spectral and curvature-related properties. Extensive experiments on both homophilic and heterophilic graphs confirm consistent performance improvements over existing rewiring strategies.


\section{GenAI Usage Disclosure}

This paper includes text that was revised and refined using generative AI tools (ChatGPT), to improve clarity and correct potential grammatical errors.  All scientific ideas, experiments, and contributions were conceived and written by the authors.

\vspace{.3cm}
\noindent \textbf{Acknowledgment.} F.M. acknowledges the support of the Innov4-ePiK project managed by the French National Research Agency under the 4th PIA, integrated into France2030 (ANR-23-RHUS-0002).

\bibliographystyle{ACM-Reference-Format}
\bibliography{ref}


\end{document}